\documentclass[sigconf]{acmart}

\renewcommand\footnotetextcopyrightpermission[1]{} 
\AtBeginDocument{%
  }

\usepackage{amsmath, amsfonts}
\usepackage{algorithmic}
\usepackage{lineno}
\usepackage{graphicx}
\usepackage{caption}
\usepackage{textcomp}
\usepackage{xcolor} 
\usepackage{booktabs}
\usepackage{multirow}
\usepackage{algorithm}
\usepackage{algorithmic}
\usepackage{subfiles}
\hypersetup{
    colorlinks=true,
    linkcolor=blue,
    urlcolor=blue,
    citecolor=blue
}
\usepackage{enumitem}

\begin{document}

\title{Rethinking Traffic Flow Forecasting: From Transition to Generatation}



\author{Shijiao Li$^{\ast}$, Zhipeng Ma$^{\ast}$, Huajun He$^{\ddagger}$, Haiyue Chen$^{\ddagger}$}

\affiliation{%
  \institution{Southwest Jiaotong University}
  \city{Chengdu}
  \state{Sichuan}
  \country{China}
}



\begin{abstract}
Traffic flow prediction plays an important role in Intelligent Transportation Systems in traffic management and urban planning. There have been extensive successful works in this area. However, these approaches focus only on modelling the flow transition and ignore the flow generation process, which manifests itself in two ways: (i) The models are based on Markovian assumptions, ignoring the multi-periodicity of the flow generation in nodes. (ii) The same structure is designed to encode both the transition and generation processes, ignoring the differences between them. To address these problems, we propose an \textbf{E}ffective \textbf{M}ulti-\textbf{B}ranch \textbf{S}imilarity Trans\textbf{former} for Traffic Flow Prediction, namely \textbf{EMBSFormer}. Through data analysis, we find that the factors affecting traffic flow include node-level traffic generation and graph-level traffic transition, which describe the multi-periodicity and interaction pattern of nodes, respectively. Specifically, to capture traffic generation patterns, we propose a similarity analysis module that supports multi-branch encoding to dynamically expand significant cycles. For traffic transition, we employ a temporal and spatial self-attention mechanism to maintain global node interactions, and use GNN and time conv to model local node interactions, respectively. Model performance is evaluated on three real-world datasets on both long-term and short-term prediction tasks. Experimental results show that EMBSFormer outperforms baselines on both tasks. Moreover, compared to models based on flow transition modelling (e.g. GMAN, 513k), the variant of EMBSFormer(93K) only uses 18\% of the parameters, achieving the same performance. Our code is available at Anonymous Github\footnote{https://github.com/shijiaoli2021/EMBSFormer}.
\end{abstract}


\maketitle

\section{Introduction}
Traffic flow prediction is a fundamental problem in ITS and is used to support various applications such as traffic control\cite{papageorgiou2003review}, route planning\cite{bast2016route}, and urban path routing\cite{yeh1999urban}.

\begin{figure}[!th]
\centerline{\includegraphics[scale=0.21]{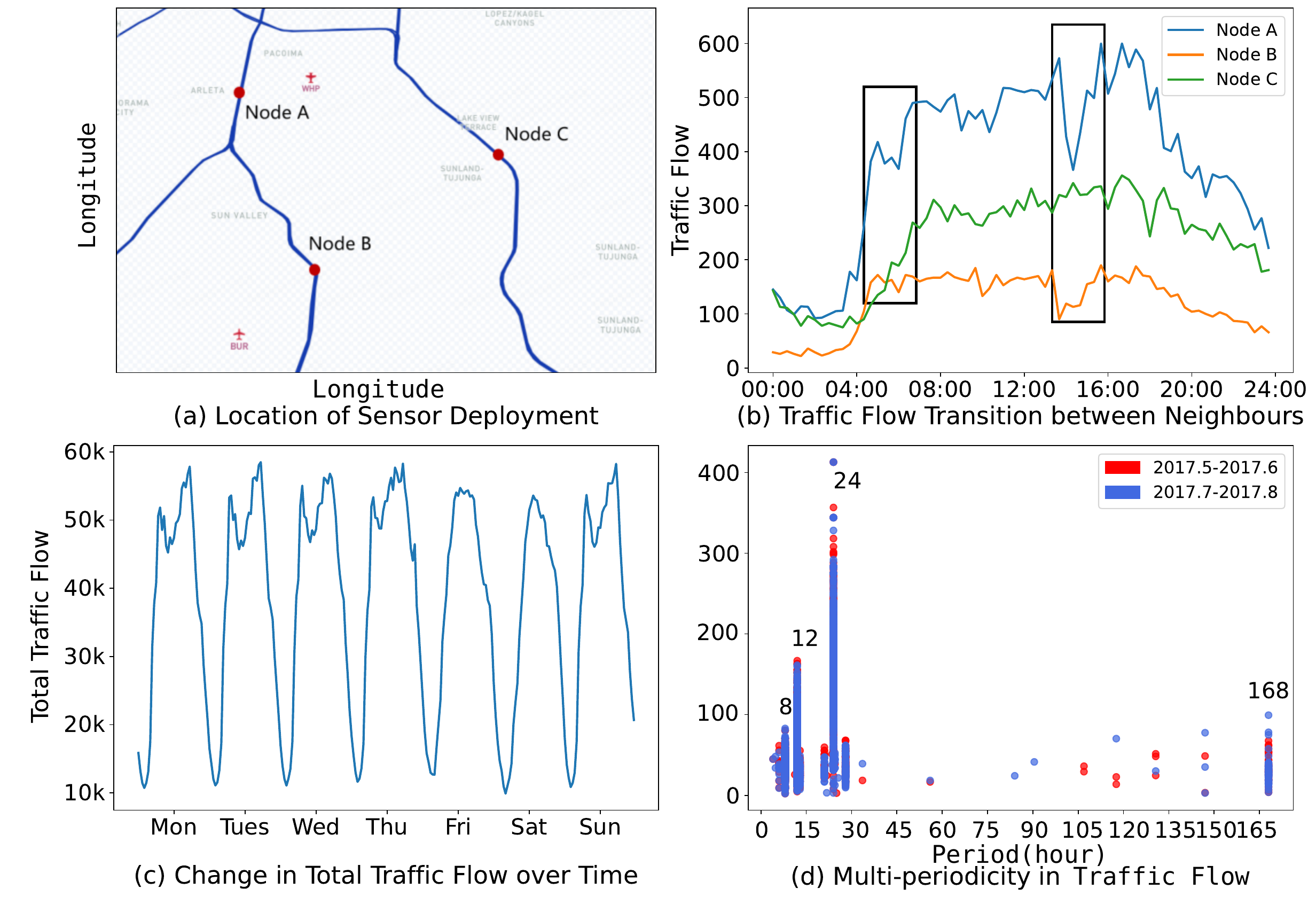}}
    \caption{Generation and Transition of Traffic Flow.}
    \label{Introduction}
    \vspace{-8pt}
\end{figure}

In past works, research community have focused on how to capture traffic transition patterns between nodes. Figure~\ref{Introduction} (a) and (c) illustrate this dynamic relationships in terms of spatial heterogeneity and temporal heterogeneity, respectively. Spatial heterogeneity is evident in the stronger correlation between neighboring nodes A and B compared to node A and node C. Additionally, the correlation level between node A and node B changes over time, reflecting temporal heterogeneity. Numerous  methods have been proposed to model these complex dynamic relationships. Initially, CNNs and RNNs were employed to capture the these pattern, as demonstrated by ST-ResNet \cite{zhang2017deep} and DMVST-Net \cite{yao2018deep}. With the advancement of GNNs, modeling unstructured spatial neighborhoods became mainstream, with representative works including ASTGCN \cite{guo2019attention} and STSGCN \cite{song2020spatial}. More recently, models such as STTN \cite{xu2020spatial}, ASTGNN \cite{guo2021learning}, and SSTBAN \cite{guo2023self}, which focus on the self-attention mechanism, have further expand GNNs to capture traffic transition pattern through implicit semantic neighborhoods.

However, previous works have overlooked the important fact that urban total traffic flows are not constant. Figure~\ref{Introduction} (c) shows the total traffic flow in the city over time, which is the sum of all node flows. It indicates that traffic flows will periodically appear and disappear, called traffic flow generation. Two problems limit the modeling performance. (i) The models are based on Markovian assumptions, ignoring the multi-periodicity of the flow generation in nodes. The Markov assumption simplifies sequence modeling by assuming that the state of the current step is only related to the state of the previous step. However, Fourier analysis of the traffic flow data, as shown in Figure~\ref{Introduction} (d), reveals a significant multi-periodic pattern. This indicates that the traffic flow at the current step is influenced not only by the previous step but also by the superposition of several significant periods. (ii) The same structure is designed to encode both the transition and generation processes, disregarding the differences between them. Complex dynamic interaction dominate flow transition modeling, while flow generation focuses more on multi-periodic temporal patterns. Previous work\cite{zhang2017deep, guo2019attention, li2021adaptive} has attempted to capture both patterns simultaneously using complex model structures and larger parameter scales. However, these approachs not only makes training more difficult to converge but also negatively impacts the model's generalization. Recent studies\cite{thompson2021deep, guha2024diminishing} have shown that the marginal gains from increasing complexity are diminishing.

To sovel above problem, we propose an efficient multi-branch similarity transformer for traffic flow prediction, named EMBSFormer. EMBSFormer consists of a stacked flow transition modules and a parallel multi-period flow generation module, designed to capture the pattern of flow transtion and the flow generation, respectively. Specifically, in the stacked flow transition module, we design spatial and temporal attention mechanisms to capture the spatial and temporal heterogeneity of traffic transitions.  Moreover, we propose a scalable multi-period flow generation module, which contains multiple flow generation blocks corresponding to the top-k significant periods. The traffic generation module and the traffic transition module are integrated using cross-attention to predicte the next step flow. . This approach significantly improves training and inference efficiency without degrading model performance. The main contributions of this paper are as follows:

\begin{itemize}[leftmargin=*]
\item We propose a new research paradigm in which traffic flow patterns are considered as a combination of flow generation and flow transition. Furthermore, a model named EMBSFormer is implemented based on this insight.
\item Specifically, we propose a flow transition module to capture transition pattern in dynamic spatio-temporal neighbourhoods. And, the multi-period flow generation module is designed to model  generation process.
\item Three real-world datasets is used on short-term and long-term traffic flow prediction tasks. The experimental results show that EMBSFormer outperforms the baselines on all prediction tasks and more efficient.
\end{itemize}

\vspace{-20pt}
\section{RELATED WORK}
\subsection{Deep Learning in Traffic Flow Prediction}
Many deep learning research efforts focus on solving traffic flow prediction, which can be classified into three categories: structured relationship modeling, neighborhood relationship modeling, and global relationship modeling. Convolutional neural network-based approaches are representative of structured relationship modeling. Due to the convolutional kernel's high efficiency in capturing locally dependent relationships within the receptive field, it is widely used for grid-based spatial relationship modeling. Also, temporal convolution and recurrent neural networks have been used to model the time dependence. This line is represented by \cite{zhang2017deep}, \cite{yao2018deep}, \cite{shi2015convolutional} and \cite{guo2019deep}. However, the transportation system consists of a road network, which is essentially a graph structure. Modeling it using a grid-based data structure results in information loss. Therefore, the use of graph neural networks to capture unstructured spatial topology has become mainstream. DCRNN \cite{li2017diffusion} uses a fully convolutional design, combining diffusion convolution for spatial dependencies and recurrent neural network for temporal modeling. AGCRN\cite{bai2020adaptive} develops a novel adaptive matrix and learns it through node embedding, and the model can precisely capture the hidden spatial dependency in the data. ASTGCN maintains a multi-branch structure to capture periodic information and utilizes attention mechanisms to capture spatial heterogeneity and recent temporal heterogeneity. TrendGCN \cite{jiang2023enhancing} extends the flexibility of GCNs by employing a dynamic graph convolutional GRU to account for heterogeneous spatio-temporal convolutions. While graph neural networks can flexibly capture unstructured neighborhood relationships, their receptive field can only be expanded by stacking network layers. To better model dependencies from a global perspective, attention mechanisms have been proposed. ASTGNN \cite{guo2021learning} and TFormer \cite{yan2021learning} employ a Transformer-like encoder-decoder structure to capture global spatio-temporal dependencies through self-attention and trans-attention, achieving remarkable results. STTN \cite{xu2020spatial} utilizes spatial and temporal self-attention to capture spatial and temporal dependencies, while GMAN \cite{zheng2020gman} designs a gated fusion to adaptively fuse information extracted from spatial and temporal attention mechanisms for prediction. DSTAGNN \cite{lan2022dstagnn} combines the advantages of both and captures explicit and implicit spatio-temporal dependencies using temporal and spatial self-attention mechanisms in spatial neighborhood matrix and an adaptive neighborhood matrix. Although these methods put significant effort into modeling spatial heterogeneity and temporal heterogeneity, they focus too heavily on capturing the flow transition pattern while overlooking the flow generation pattern. In contrast, we divide the traffic flow pattern into flow generation and flow transition stages, and we use different modules to model them according to their distinct characteristics.

\subsection{Transformer} 
Transformer \cite{vaswani2017attention} is an encoder-decoder network structure based on an attention mechanism. It transfers information between the encoder and decoder through cross-attention. Benefiting from the Transformer's excellent learning ability and flexible modeling capability, it has found wide applications in natural language processing \cite{devlin2018bert}, computer vision \cite{liu2021swin}, and sequence modeling \cite{zhang2022crossformer}. Unlike the general encoder-decoder structure of the Transformer, we construct a multi-branch encoder-only model, which achieves historical memory lookup through similarity attention. Compared with the encoder-decoder structure, it offers higher inference efficiency.

\section{Preliminaries}

\noindent

\textbf{Definitions 1} (Traffic Flow Graph). A traffic flow graph is define as $\mathcal{G}=\left(\mathcal{V},\mathcal{E},\mathcal{A} \right)$, where $\mathcal{V}$ is a set of nodes containing $N$ spatial locations and $\mathcal{E}$ represents the set of edges between neighboring nodes. $\mathcal{A}\in \mathbb{R}^{N \times N}$ denotes the adjacency matrix, which is a binary matrix where an entry is 1 if two nodes are adjacent and 0 otherwise.

\textbf{Definitions 2} (Traffic Flow Data). We use $X_t$ to represent the current traffic flow at all nodes in time $t$, denoted as $X_t \in \mathbb{R}^{N \times F}$. Furthermore, the traffic flow data for the past $m$
time steps is denoted as $X_{t-m+1:t}\in \mathbb{R}^{m \times N \times F}=(X_{t-m+1},X_{t-m+2},\dots,X_{t})$.

\textbf{Problem Definitions}. 
Given the historical traffic flow data $X_{t-m+1:t}\in \mathbb{R}^{m \times N \times F}$ up to step $t$, our goal is to learn a function that predicts the traffic flow at all nodes for the next $n$ steps $X_{t+1:t+n} \in \mathbb{R}^{n \times N \times F}$.

\section{Methodology}

\subsection{Overview}



\begin{figure*}
	\centering
	\includegraphics[width=1\linewidth]{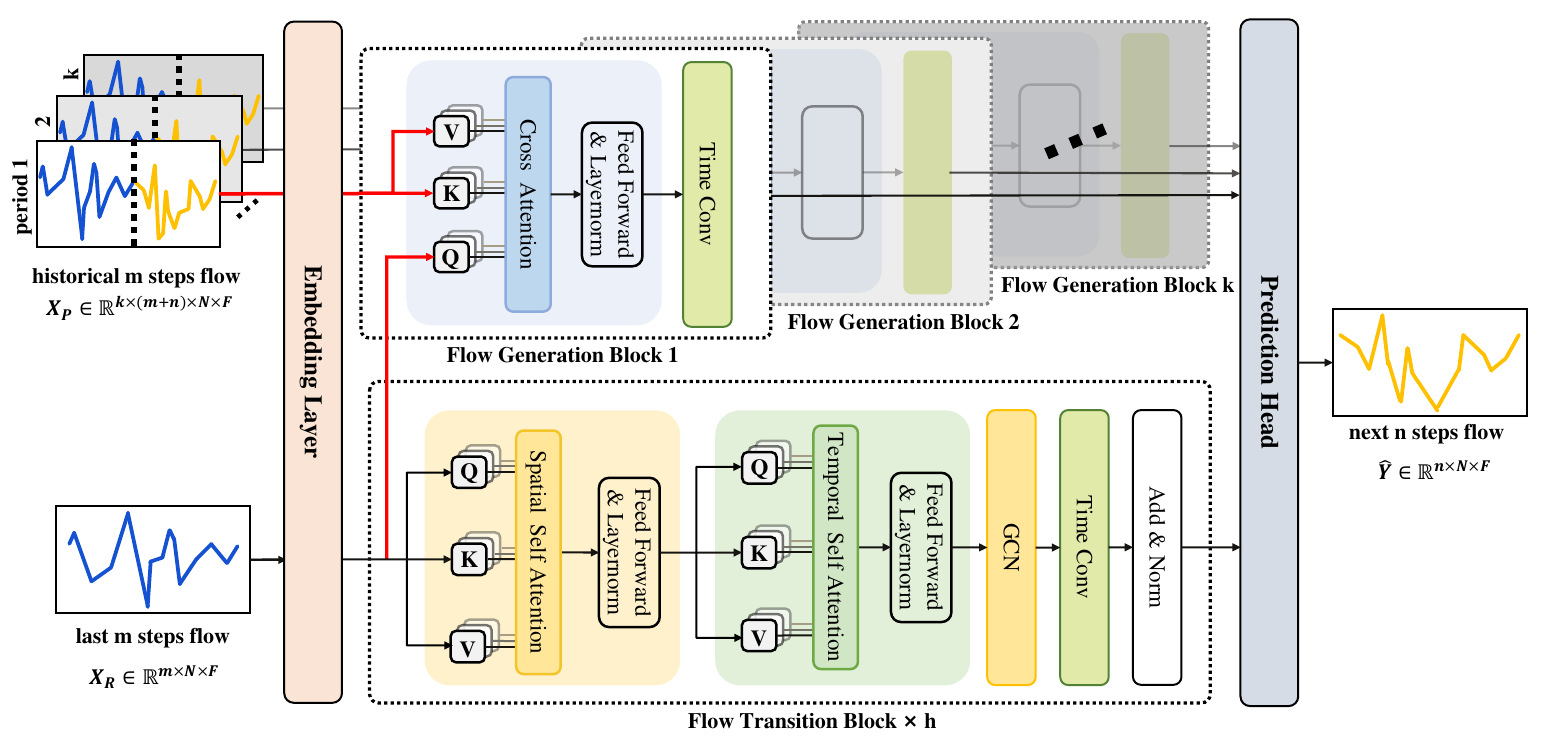}
	\caption{Framework of EMBSFormer.}
	\label{Framework}
\end{figure*}

The framework of EMBSFormer is shown in Fig. \ref{Framework}, which consists of two main branches: flow generation and flow transition.

. Initially, we extract historical data and divide it into two parts: recent data and multi-period historical data. These data are then fed into the embedding layer for feature mapping.

In the flow encoding layer, we employ different learning strategies for short-term and long-term historical data. 

On one hand, we propose a Parallel Multi-Period Flow Generation Module focused on capturing flow generation patterns. 

On the other hand, we devise a stacked flow transition module to learn the propagation relationships between traffic flows. 

Finally, the outputs of each module are fused through the output layer to obtain the prediction result.

\subsection{Multi-period Data Preparation.}
\begin{figure}[htbp]
    \centering
    \includegraphics[width=1.05\linewidth]{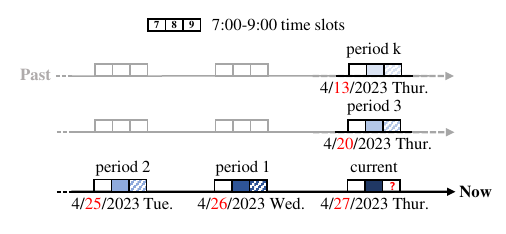}
    \caption{Multi-period Data Preparation.}
    \label{Time series}
\end{figure}

We divide the historical traffic flow into two parts: the short-term history (recent data) and the long-term historical flow (multi-period data). The recent historical flow represents the time series of the specified length $m$ before the time $t$, which has a continuous relationship with the sequence to be predicted in data collection. For long-term historical traffic, we simultaneously sample $m+n$ sequence data from historical data with a period of $T$ for similarity analysis. For example, as shown in Fig. \ref{Time series}, in order to predict the traffic from 8:00 to 9:00 on April 27, 2023, we partition the historical data into segments based on temporal duration. We construct short-term historical data from 7:00 to 8:00 on April 27, 2023, and long-term historical data by aggregating data in intervals of days and weeks, specifically from 7:00 to 9:00, such as 7:00 to 9:00 on April 26, 2023, and 7:00 to 9:00 on April 20, 2023. The mathematical definition of historical traffic input is as follows:


\begin{itemize}[leftmargin=*]
	\item \textbf{The recent flow.} \\$X_R
	\in \mathbb{R}^{N \times T_{in} \times F}$, which is directly adjacent to the traffic time series to be predicted and directly impacts future predicted sequences.
        \item\textbf{The period flow.}
        \\$X_P\in \mathbb{R}^{K \times N \times (T_{in}+T_{out}) \times F}$,  where $T$ represents the interval of period and $T \geq m+n$, $k$ is the num of period. 
\end{itemize}

\vspace{-10pt}

\subsection{Embedding Layer}
The data embedding layer aims to map the input feature dimensions to a high-dimensional representation. In practice, the original input \(X \in [X_R, X_P]\) is embedded into \(E_{R} \in \mathbb{R}^{m \times N \times d_e}\) and \(E_{P} \in \mathbb{R}^{(m+n) \times N \times d_e}\), respectively.

Referring to previous work \cite{zhang2017deep}, we extract multiple temporal features to enhance model performance. We augment the input by incorporating timestamps and specifying the minute, hour, and day to facilitate the model in capturing periodic traffic flow patterns. Specifically, the minute of the day (\([0, 1443]\)) and the day of the week (\([0, 6]\)) are embedded. Additionally, we annotate holidays (\(0\) or \(1\)) at each time step to account for the influence of holidays on predictions. These features are embedded into \(h\) dimensions, represented as \(X^{min}\), \(X^{week}\), and \(X^{holiday}\). Furthermore, to express the order relationship of time series, we introduce a trigonometric function positional encoding, embedded as \(X^{pos}\), to the original data. The data embedding can be represented as follows:

\begin{equation}
	E =X^{data}+X^{min}+X^{week}+X^{holiday}+X^{pos}
	\label{embedding layer}
\end{equation}

where $E=\left[E_{R}, E_{P}\right]$, they represent recent data and multi-period historical data, respectively. 

\subsection{Flow Encoding Layer}
This layer is the core of the model, consisting of the stacked flow transition moudle and the parallel multi-period flow generation Module. In stacked flow transition moudle, the spatio-temporal self-attention mechanism and graph convolution are employed to analyze the direct propagation relationships of traffic flow in the spatio-temporal topology. In parallel multi-period flow generation Module, a similarity attention mechanism is introduced to analyze the similarity between long-term and recent histories, providing guidance for predictions. The details are as follows.
\subsubsection{Flow Transition Block}
We employ spatial-temporal encoding modules on short-term historical flows based on the self-attention mechanism to capture spatio-temporal dependencies. Its core mainly includes three parts. At first, the spatial self-attention mechanism captures different points' dependencies simultaneously. Secondly, we employ a temporal self-attention mechanism to capture recent historical traffic's temporal dependence and dynamics. Thirdly, we employ a graph convolutional neural network to express adjacency information.

\textbf{spatio-temporal neighbourhood attention.}
    We design a spatial self-attention mechanism to capture the dynamic dependence of space. Firstly, we map the input vectors in the spatial dimension to obtain $Q_S$, $K_S$, $V_S\in \mathbb{R}^{m\times N\times d_s}$ of space as query, key, value respectively through the trainable parameters $W_S^Q$, $W_S^K$, $W_S^V\in \mathcal{R}^{d_e\times d_s }$ as follows:

\begin{equation}
\begin{aligned}
Q^S= W^S_Q E_{R}+b^S_Q\\
K^S= W^S_K E_{R}+b^S_K\\
V^S= W^S_V E_{R}+b^S_V
\end{aligned}
\end{equation}

Where $b_q$,$b_k$,$b_v\in R^{d_s}$ represent bias. Then, through the operation of the self-attention mechanism, we can obtain the dependency relationship $A_S^r\in R^{\left(N\times N\right)}$ at different time points and in different spaces:

Finally, we obtain the output of the spatial self-attention mechanism by matrix multiplication between the dependency and the value, denoted as $SSA$:
\begin{equation}
	E^{S}_R =Softmax\left(\frac{Q^S K^S}{\sqrt{d_s}}\right) V^S
\end{equation}

\textbf{The Temporal Self-Attention.} In the temporal self-attention mechanism,  we map the input vectors in the temporal dimension to obtain $Q^T$, $K^T$, $V^T\in R^{N\times t\times d_t }$ respectively through the trainable parameters $W_Q^T$,$W_K^T$, $W_V^T\in R^{d_s\times d_t }$:

\begin{equation}
\begin{aligned}
Q^T= W^T_Q E^{S}_R+b_T^Q\\
K^T= W^T_K E^{S}_R+b_T^K\\
V^T= W^T_V E^{S}_R+b_T^V
\end{aligned}
\end{equation}

Among them, $b_q$, $b_k$, $b_v\in R^{h^{'}}$ represent the bias, and $T$ represents the dimension of transposing the time series and the point. Then, we get the dependency relationship $A_S^r\in R^{\left(N\times m\times m\right)}$ at different points and at different times through the operation of the self-attention mechanism:

\begin{equation}
	E^T_R=Softmax\left(\frac{Q^T K^T}{\sqrt{d_t}}\right)V^T
\end{equation}

\textbf{Spatio-temporal neighbourhood convolution.}

In the traffic flow prediction, We introduce a graph convolutional neural network with Chebyshev polynomials as the core to represent the topological relationship of different points in the traffic network, facilitating the model in capturing the adjacency flow relationships between locations during the learning process.

Firstly, we obtain the Laplacian matrix through the adjacency matrix and the degree matrix to represent the adjacency relationship $L=D-A$ of the nodes in the graph. Where D represents the degree matrix, $D\in \mathbb{R}^{N\times N}$ represents the adjacency matrix, and its regularization expression is $L= I_N- D^{\frac{-1}{2}}AD^{\frac{-1}{2}}$. The eigenvalue decomposition of the Laplacian matrix is expressed as $L=U\Lambda U^T$, where $\Lambda =diag\left(\lambda_1,\lambda_2…\lambda_N\right)$. According to the definition of convolution, we use $U^Tx$ to $x$ Fourier transform and instead regard $U$ as the inverse Fourier transform. Then, the spatial graph convolution transform is expressed as follows:

\begin{equation}
	g_{\theta}*\mathcal{G}x=U\left(U^T g_{\theta} U^T x\right)=Ug_{\theta} U^Tx
\end{equation}

Where $g_{\theta}$ represents the function of the eigenvalue of $\Lambda$, and the Chebyshev polynomial is quoted here to represent the function of the eigenvalue as follows:

\begin{equation}
	g_{\theta}= \sum_{k=0}^{K-1}{\theta_k T_k \tilde{\Lambda} x}
\end{equation}

Where $\tilde{\Lambda}= \frac{2\Lambda}{\lambda_{max}}-1$, $T_k$ represents the Chebyshev polynomial $T_k \left(x\right)=2\times T_{k-1} \left(x\right)-T_{k-2} \left(x\right)$, $T_0 {x}=1$, $T_1 {x}=x$. Substituting the above formula into the simplification to obtain the graph convolution expression is as follows:

\begin{equation}
	gcn\left(x\right) = ReLu\left(g_\theta*\mathcal{G}x\right)=ReLu\left(\sum_{i=0}^{K-1}{\theta_iT_i\tilde{L}}\right)
\end{equation}

And finally get the graph convolution output $ReLu\left(g_{\theta}*\mathcal{G}x\right)$ through the $ReLu$ activation layer.

After we perform graph convolution to capture the adjacency information of each point, we finally convolve the traffic information in the time dimension to update the kernel to capture the sequence characteristics of each point at different times and add a residual structure to it, which is regarded as $ASTG\left(X_r \right)$, expressed as follows:
\begin{equation}
	\hat{Y}_R = ASTG\left(X_r\right) = conv_F \left(conv_t\left(gcn\left(E_R^T\right)\right)\right)\in \mathbb{R}^{n\times N}
	\label{recent block eq}
\end{equation}

Among them, $conv_F$ represents the convolution operation on the feature, $conv_t$ represents the convolution operation on time, and $conv_r$ represents the convolution operation on the initial input feature.

\begin{figure}[th]
	\centering
	\includegraphics[width=0.9\linewidth]{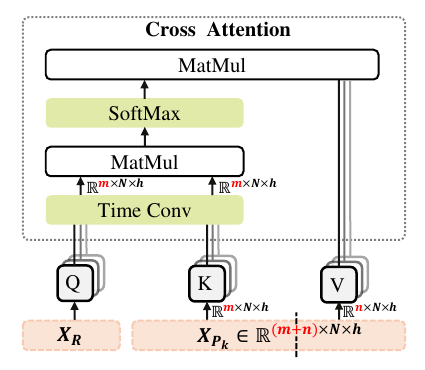}
	\caption{The Multi-period lookup Attention. }
	\label{similarity attention}
 \vspace{-0.1cm}
\end{figure}

\subsubsection{Flow Generation Block}
To tackle the muti-periodic patterns in historical traffic flow, we introduce a long-short-term history similarity attention mechanism specifically for the long-term historical flow. This mechanism assesses the similarity between traffic flow in the long-term historical cycle and short-term historical traffic flow. The output is obtained through temporal convolution, enhancing traffic prediction and augmenting the accuracy of model predictions.


\textbf{Similarity Attention.}
Drawing inspiration from the periodic nature of flow patterns, we obviate the necessity for managing an additional self-supervised or alternative model to address heterogeneity and dependencies in operational modes. Addressing both dependency issues can be achieved by simply focusing on and studying the similarity of historical traffic flow. As shown in Fig. \ref{similarity attention}, firstly, We split the long-term historical data into long-term historical input and pseudo-future vectors based on the input and prediction lengths, as memory bank $E_P \in \mathbb{R}^{K \times N \times (T_{in}+T_{out}) \times F}$

$E_P^i(in) \in \mathbb{R}^{ N \times T_{in} \times F}$

$E_P^i(out) \in \mathbb{R}^{ N \times T_{out} \times F}$

. k denotes the number of significant cycles.

Secondly, we separately map recent history, long-term history, and pseudo-future data into a high-dimensional space to obtain $Q_S^e$, $K_S^e$, $V_S^e\in R^ {\left(m\times N\times h^{'} \right)}$ of space as query, key, value respectively through the trainable parameters $W_S^q$, $W_S^k$, $W_S^v\in R^{\left(h\times h^{'} \right)}$ as follows:

\begin{equation}
\begin{aligned}
Q^M= W^M_Q E_P+b^M_Q\\
K^M= W^M_K E_P+b^M_K\\
V^M= W^M_V E_P+b^M_V
\end{aligned}
\end{equation}

Next, we calculate the similarity among historical data and leverage the similarity, along with the pseudo-future data, to predict traffic flow patterns. However, due to the inconsistency in input and output lengths, we initially utilize time convolution to transform the input length to match the output length  $\left(m, N,h^{'}\right)$→$\left(n, N,h^{'}\right)$ as follows:
%
\begin{equation}
\begin{aligned}
Q^M= W^M_Q E_P+b^M_Q\\
K^M= W^M_K E_P+b^M_K\\
V^M= W^M_V E_P+b^M_V
\end{aligned}
\end{equation}

$$
CQ^{'}_r=conv_r\left(T\left(Q_S^e\right)\right)
$$

\begin{equation}
	CK^{'}_e=conv_w\left(T\left(V_S^e\right)\right)\in R^{N\times n\times h^{'}}
\end{equation}

The $T$ function represents the exchange time and point dimensions order. Then, We calculate the time series similarity $A_e^r\in R^{N\times n\times n}$ between the long-term historical cycle data and the short-term historical cycle data as follows:
\begin{equation}
	A^e_S = 
\end{equation}

Finally, we multiply the similarity relationship with $V_e^r$ to obtain the output of the long-term and short-term historical similarity attention mechanism, which is recorded as $ASR_e\in R^{n\times N\times h^{'}}$:

\begin{equation}
	ASR_e = T^{'}softmax\left(\frac{Q^e_S K^e_S}{\sqrt{h^{'}}}\right)V^e_S)
\end{equation}
where $T^{'}$ represents the inverse transformation of $T$.

\textbf{Temporal Convolution.}
After the output through the similarity attention mechanism, it is then convolved on the time series further to capture the valuable information on the long history cycle, and finally obtain the output $\hat{Y}_e\in R^{n×N}$. We employ sum normalization instead of weighted summation to highlight the similarities among historical features better and dramatically reduce the number of parameters when increasing the number of historical inputs.

\begin{equation}
	\hat{Y}_p=Norm\left(\sum_{i=1}^{q}{\left(Conv_c\left(Conv_t\left(ASR_e\right)\right)\right)}\right)
	\label{similarity block eq}
\end{equation}

Among them, $Conv_t$, $Conv_c$ means $(1,1)$, $(1,h^{'})$ convolution operation on time series and hidden features, Norm means normalization operation.

\subsection{Prediction Head}
We obtain the final prediction output $\hat{Y} \in R^{n\times N}$ by learnable weight, respectively, as follows:

\begin{equation}
	\hat{Y} = W_r\odot\hat{Y}_r + \sum_{i=1}^{c}{W^i_p\odot\hat{Y}_p^i}
	\label{output layer eq}
\end{equation}

Where $\odot$ represents the dot product, $W_r$,$W_i\in R^{n\times N}$ are trainable weight parameters, and the final prediction result can be obtained by clicking on the three training parameters and the output of each module. Finally, we employ Mean Square Error (MSE) to optimize the model, which is widely used for regression problem:

\begin{equation}
    MSE = \frac{1}{n}\sum_{i=1}^{n}\left(\hat{y}_i-y_i\right)^2
\end{equation}

\section{EXPERIMENTS}

\subsection{Datasets}
We evaluate our approach on three real-world datasets: PEMS04, PEMS07, and PEMS08. The data quality of PEMS03 is poor, so removing it. These datasets are collected from the traffic database of the California freeway network, and detailed information is provided in the Table~\ref{Statistics of Datasets}. 

\begin{table}[th]
        \setlength{\abovecaptionskip}{0.3cm}
	\caption{Statistics of Datasets.}
	\label{Statistics of Datasets}
	\begin{center}
             \setlength{\tabcolsep}{1mm}{
		\renewcommand\arraystretch{1.2}
		\begin{tabular}{c|c c c c c}
			\toprule[1.6px]
			&Start date&End date&Edges&Nodes&Time interval\\
			\hline
			PEMS04&01/01/2018&02/28/2018&340&307&5min\\
			PEMS07&05/01/2017&08/31/2017&866&883&5min\\
			PEMS08&07/01/2016&08/31/2016&295&170&5min\\
			\toprule[1.6px]
		\end{tabular}}
	\end{center}
 \vspace{-0.4cm}
\end{table}

\subsection{Baselines}
We compare EMBSFormer with the following 11 baselines, which are divided into three classes:
\begin{itemize}[leftmargin=*]

    \item \textbf{Statistical Methods. } We chose VAR \cite{lu2016integrating} and SVR \cite{drucker1996support} as representatives of statistical methods. VAR (Vector Autoregression) models linear relationships among multiple traffic flow series by using each series' past values. SVR (Support Vector Regression) employs a support vector machine for regression task, aiming to fit data using a non-linear approach.
    
    \item \textbf{GNN-Based Models.} GNN-based approach focuses on modeling the unstructured correlation of spatial and temporal neighborhoods. DCRNN \cite{li2017diffusion} combines diffusion convolution for spatial dependencies and recurrent neural network for temporal modeling with encoder-decoder framework training. GWNET \cite{wu2019graph}, which organizes graph convolutional network and gated temporal convolutional networks together. ASTGCN \cite{guo2019attention} builds on DCRNN by accessing periodic information through a multi-branch structure and using attention mechanisms to capture spatial heterogeneity and recent temporal heterogeneity. AGCRN\cite{bai2020adaptive} introduces adaptive adjacency matrix to capture spatial heterogeneity more efficiently through node learning and graph generation techniques. In STGNCDE\cite{choi2022graph},  two Neural Controlled Differential Equations are designed to capture temporal and spatial dependence modeling, respectively. TrendGCN \cite{jiang2023enhancing} extends the flexibility of GCNs by employing a dynamic graph convolutional GRU to account for heterogeneous spatio-temporal convolutions

    \item \textbf{Attention-based Models.} Attention-based approach extends local unstructured correlations to global view. GMAN \cite{zheng2020gman} uses an encoder-decoder framework for prediction via transformer attention and employs spatial and temporal attention mechanisms instead of graph convolution and temporal convolution. ASTGNN \cite{guo2021learning}, designing trend-aware self-attention module and a dynamic graph convolution module to capture spatio-temporal dynamics. In SSTBAN \cite{guo2023self}, a spatio-temporal bottleneck attention mechanism is designed to efficiently encode global spatio-temporal dynamics at a low computational cost.
\end{itemize}

\subsection{Training Details \& Hyperparameters}
\subsubsection{Data Preprocessing.}
Three datasets are divided into the training set, validation set, and test set by a ratio of 6:2:2, chronologically. All data were normalized according to $x^{'}=\frac{x-mean}{std}$, where mean and std represent the mean and standard deviation of the training set.
\subsubsection{Hyperparameters.}
The code is implemented using the PyTorch 2.0.0 framework with Python version 3.8.16. The models are trained on an NVIDIA A4000 24G GPU. Furthermore, the model optimizer uses the Adam optimization algorithm and sets the learning rate to 0.001. Set the batch size to 16 and the training epoch to 100. We adjust different parameters for training on the two types of tasks: input length \( m \) and output length \( n \). For short-term prediction, we set \( m = n = 12 \); for long-term prediction, we set \( m = n = 36 \). We experiment with the following parameter settings: Chebyshev polynomial order \( K = \{2,3\} \), hidden layer dimensions \( h' = \{32, 64, 128\} \), and data embedding dimensions \( \text{embedding dim} = \{32, 64\} \). The best hyperparameters of model is also selected based on
their performance on the validation set.

\subsubsection{Evaluation Metrics}
We evaluate model performance by adopting widely used metrics: Mean Absolute Error (MAE), Root Mean Square Error (RMSE), and Mean Absolute Percentage Error (MAPE).
$$MAE=\frac{1}{n}\sum_{i=1}^{n}{\left|\hat{y}_i-y_i\right|}$$
$$RMSE=\sqrt{\frac{1}{n}\sum_{i=1}^{n}\left(\hat{y}_i-y_i\right)^2}$$
$$MAPE = \frac{100\%}{n}\sum_{i=1}^{n}\left|\frac{\hat{y}_i-y_i}{y_i}\right|$$

\begin{table*}[ht]
	\caption{Performance Comparison on Short-term Prediction.}
	\begin{center}
		\renewcommand\arraystretch{1.2}
		\begin{tabular}{c|c c c|c c c|c c c}
			\toprule[1.6px]
			\multirow{2}{*}{Model}&\multicolumn{3}{c}{PEMS04-12 steps}&\multicolumn{3}{c}{PEMS07-12 steps}&\multicolumn{3}{c}{PEMS08-12 steps}\\
			\cline{2-10}
			\multirow{2}{*}{}&MAE&RMSE&MAPE&MAE&RMSE&MAPE&MAE&RMSE&MAPE\\
			\hline
			VAR&23.75&36.66&18.09&101.20&155.14&39.69&22.32&33.83&14.47\\
			SVR&28.66&44.59&19.15&32.97&50.15&15.43&23.25&36.15&14.71\\
		\hline
            DCRNN (ICLR'18)&22.74&36.58&14.75&23.63&36.51&12.28&18.19&28.18&11.24\\
            ASTGCN (AAAI'19)&21.54&34.27&16.17&22.03&35.22&9.85&17.60&24.62&10.50\\
            GWNET (IJCAI'19)&19.36&31.72&13.30&21.22&34.12&9.08&15.06&24.86&9.51\\
            AGCRN (NIPS'20)&21.19&33.65&13.90&21.97&34.81&10.12&15.95&25.22&10.09\\
            STGNCDE (AAAI'22)&19.21&31.09&12.77&20.62&34.04&8.86&15.46&24.81&9.92\\
            TrendGCN (CIKM'23)&18.81&31.20&\textbf{12.36}&21.16&34.52&\textbf{8.82}&15.06&\underline{24.25}&9.77\\
            \hline
            
            GMAN (AAAI'20)&19.38&31.60&13.19&20.97&34.10&9.05&15.31&24.92&10.13\\     
            ASTGNN (TKDE'21)&\underline{18.68}&\underline{31.03}&12.63&\underline{20.61}&\underline{34.02}&8.86&\underline{14.78}&24.71&\underline{9.49}\\
            SSTBAN (ICDE'23)&18.88&31.24&12.98&-&-&-&15.32&24.26&10.73\\

            \hline
            \textbf{EMBSFormer}&\textbf{18.42}&\textbf{30.51}&\underline{12.43}&\textbf{20.44}&\textbf{33.13}&\underline{8.84}&\textbf{14.29}&\textbf{23.33}&\textbf{9.33}\\
		\toprule[1.6px]
		\end{tabular}
	\end{center}
	\label{short term}
\end{table*}

\begin{table}[ht]
	\caption{Performance Comparison on Long-term Prediction.}
	\begin{center}
            \setlength{\tabcolsep}{1mm}{
		\begin{tabular}{c|c c c|c c c}
			\toprule[1.6px]
			\multirow{2}{*}{Model}&\multicolumn{3}{c}{PEMS04-36 steps}&\multicolumn{3}{c}          {PEMS08-36 steps}\\
			\cline{2-7}
			\multirow{2}{*}{}&MAE&RMSE&MAPE&MAE&RMSE&MAPE\\
			\hline
			VAR&30.48&45.44&24.51&31.70&48.96&22.56\\
			SVR&27.25&43.15&22.71&26.25&53.15&21.71\\
    		\hline
                DCRNN (ICLR’18)&33.78&51.40&27.10&25.82&39.37&18.53\\
                GWNET
                (IJCAI’19) &24.71&38.17&17.67&18.57&29.06&14.06\\
                AGCRN (NIPS’20)&24.15&38.19&16.33&19.39&30.96&12.73\\
                TrendGCN (CIKM’23) &\underline{21.10}&34.84&\underline{13.88}&17.60&28.17&11.73\\
                \hline
                GMAN (AAAI’20) &22.12&52.86&16.43&\underline{17.21}&33.96&16.33\\
                SSTBAN (ICDE’23)&21.11&\underline{34.48}&14.92&17.31&\underline{27.89}&\underline{11.60}\\
                \hline
                \textbf{EMBSFormer}&\textbf{20.49}&\textbf{33.79}&\textbf{13.66}&\textbf{16.14}&\textbf{26.12}&\textbf{11.12}\\
		\toprule[1.6px]
		\end{tabular}}
	\end{center}
	\label{long term}
\end{table}

\subsection{Overall Performance}
We conducted experiments separately for short-term and long-term forecasting. Table \ref{short term} and Table \ref{long term} show the average results on two tasks. Bold scores and underline scores indicate the best and the second best, respectively. The analyses are as follows:

On the one hand, the comparison results of short-term prediction with three datasets are shown in Table \ref{short term}. We can observe the following points from the table: (1) For the time series prediction model, VAR and SVR only account for the temporal dependencies between time series and do not consider traffic flow features. They performed poorly in the baseline experiments. (2) The graph convolution-based models, which consider the dependency on the adjacency relationship between nodes, consistently outperform traditional time series forecasting methods. Moreover, among them, TrendGCN shows the better results in three datasets. (3) The self-attention models that adopt self-attention to analyze spatiotemporal dependencies acquire better results than GNN-based models, and ASTGNN shows the best among them. (4) the EMBSFormer outperforms all baselines, especially on the PEMS08 dataset, exhibiting improvements of MAE(3.32\%), MAPE(1.69\%), and RMSE(3.79\%) compared to ASTGNN on the PEMS08 dataset, which unequivocally highlights the exceptional efficacy of EMBSFormer in short-term forecasting. 

\begin{figure}[h]
	\centerline{\includegraphics[scale=0.285]{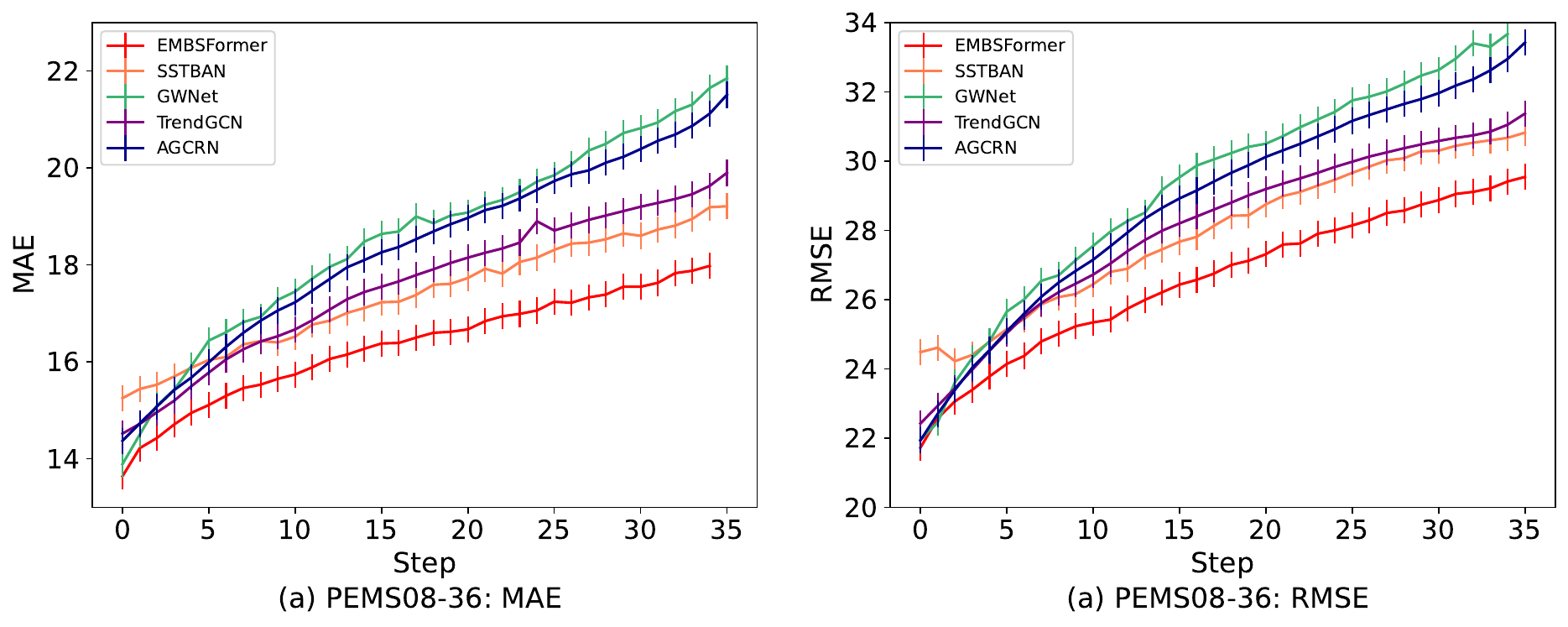}}
	\caption{Long-term Prediction Performance with Increasing Steps.}
	\label{step_prediction}
\end{figure}

On the other hand, the comparison results of long-term prediction with PEMS04 and PEMS08 are shown in Table \ref{long term}. In this study, we conducted modelling experiments to predict the next three hours. (1) Among all baselines, SSTBAN almost performs the best, addressing long-term forecasting challenges using a self-supervised architecture. (2) The prediction accuracy decreases as the forecasting horizon extends, as observed in SSTBAN, which exhibits better performance at 12 steps. (3) In the context of long-term forecasting tasks, EMBSFormer has been exhibited superior performance. Notably, in long-term forecasting, EMBSFormer's superiority becomes more pronounced. Compared to the second best, it achieves improvements of 2.89\%, 1.59\%, and 2.00\% for PEMS04, and 6.22\%, 4.14\%, and 6.34\% for PEMS08 in terms of RMSE, MAPE, and MAE at 36 steps. Figure \ref{step_prediction} shows the changing of prediction errors on the 36-step prediction task on the PEMS08 dataset. We find that the advantages of our EMBSFormer become obvious as the prediction span extends. The improvement of these results robustly demonstrates the ability of EMBSFormer to address long-term forecasting challenges. Observing EMBSFormer's prediction performance, we note that it remains largely unaffected by the increase in forecasting horizon. This resilience is attributed to the similar attention mechanism, which can perform similarity analysis at any length, and the module's efficiency in parameter usage.

Moreover, compared with ASTGCN, which also considers multi-period, EMBSFormer outperforms ASTGCN significantly. It is noteworthy that ASTGCN employs a more complex approach to handling long-term history. ASTGCN utilizes multiple layers of attention mechanisms and graph convolution operations, while EMBSFormer employs only a single layer of similar attention mechanisms.

In conclusion, the experiments thoroughly demonstrate EMBSFormer's outstanding performance in both short-term and long-term forecasting. Furthermore, we will illustrate the effectiveness of each module in the experiments through ablation studies.

\subsection{Ablation Study}

\begin{figure*}[t]
	\centering
	\includegraphics[width=0.95\linewidth]{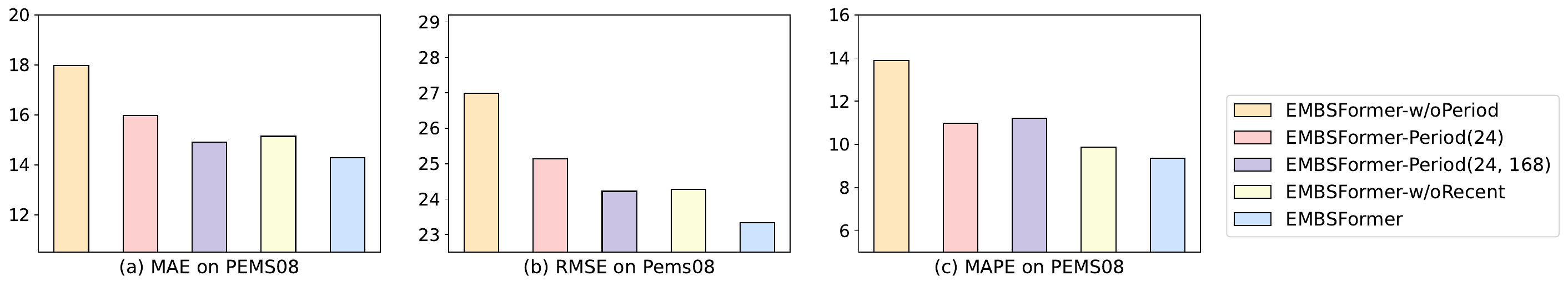}
	\caption{Ablation Study.}
	\label{ablation study}
    \vspace{-0.4cm}
\end{figure*}

\begin{figure*}[h]
	\centering
	\includegraphics[scale=0.3]{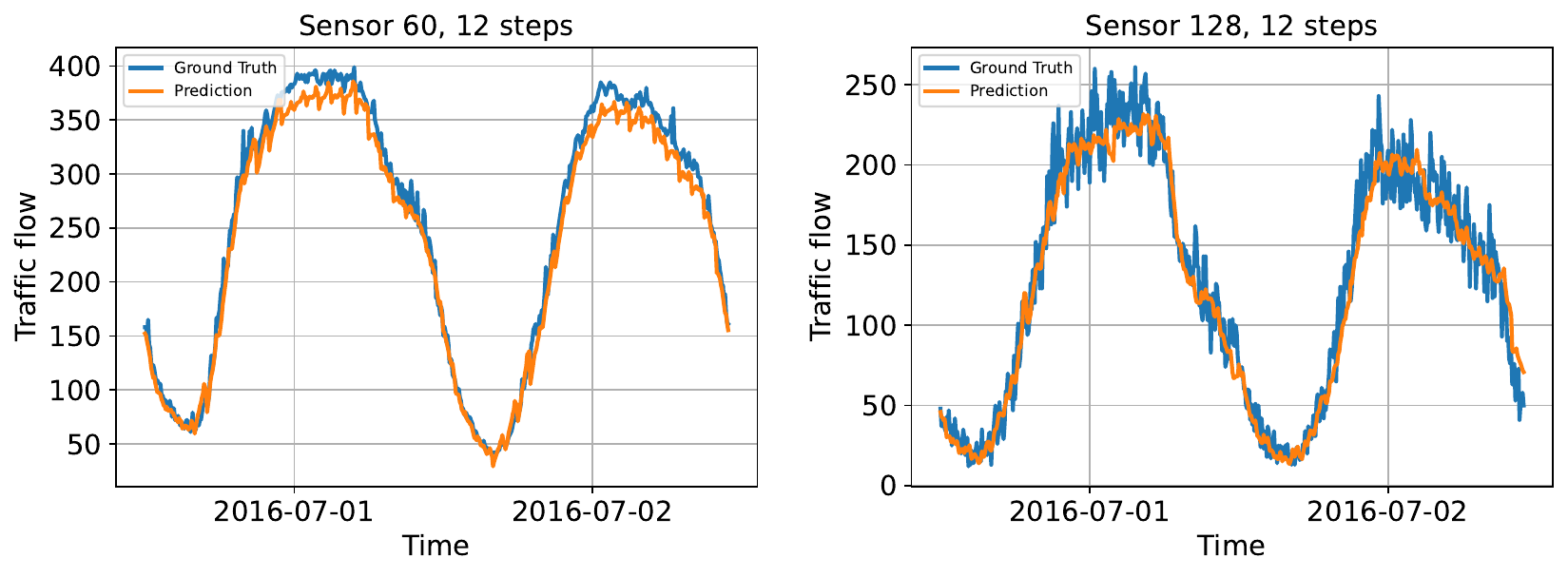}
	\includegraphics[scale=0.3]{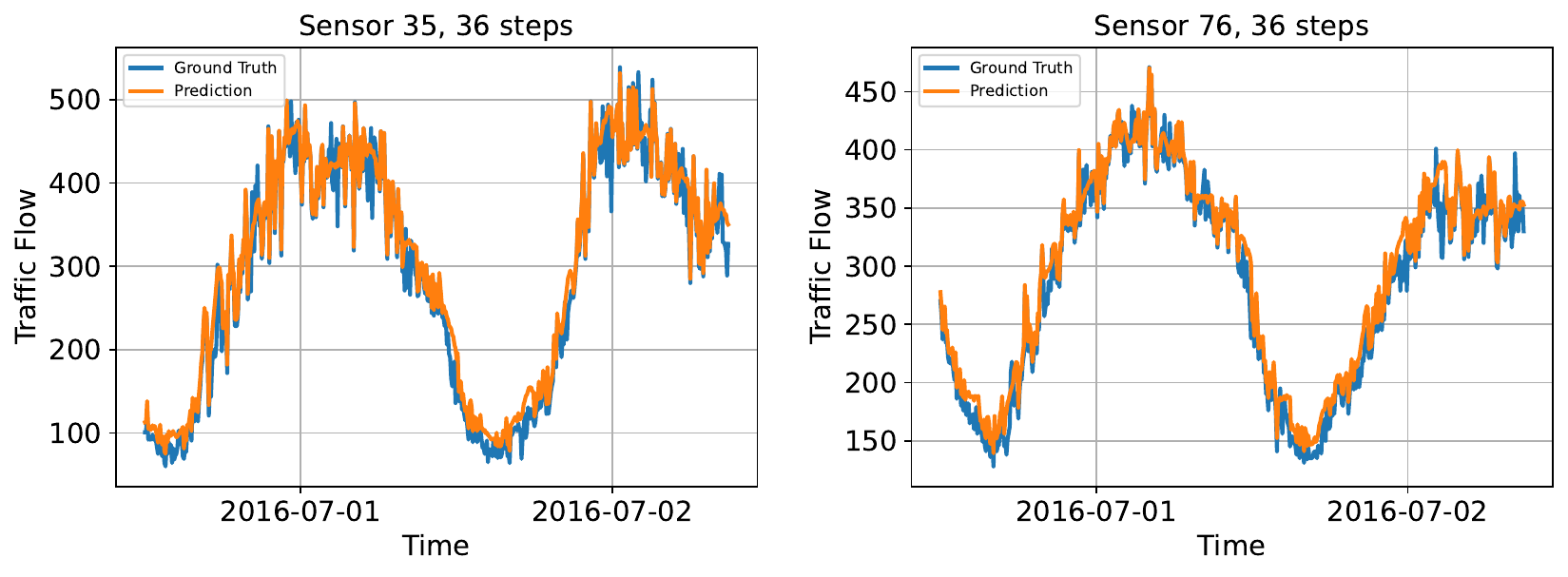}
	\caption{Case Study.}
	\label{predition curve}
\end{figure*}
To assess the model's efficacy, we conducted ablation experiments, wherein specific components were removed from the model to ascertain their impact on the overall performance: (1) \textbf{EMBSFormer}. The model samples long-term historical information with 8, 12, 24 and 168 hours on the similarity analysis module. (2) \textbf{EMBSFormer-w/o Period}. It excludes the Period module and focuses solely on the analysis of recent attention. (3) \textbf{EMBSFormer-Period (24)}. It samples only long-term historical information 24 hours on the similarity analysis module. (4) \textbf{EMBSFormer-Period (24,168)}. It samples long-term historical information with 24 and 168 hours on the similarity analysis module. (5) \textbf{EMBSFormer-w/o Recent}. It excludes the recent module in the recent block. Figure \ref{ablation study} shows the ablation study on short-term prediction. 

Based on the results, we can conclude the following: 1. The significance of efficient utilization of multi-cycle information: From EMBSFormer-Period(24), EMBSFormer-Period(24, 168), EMBSFormer-Period(8, 12, 24, 168), it is evident that the model prediction error diminishes as the sampling of multi-period information for similarity analysis increases, which indicates that the optimal utilization of long-term information can enhance the fit. 2. Multi-periodic analysis predicts the overall direction of traffic flow. In contrast, the near-term trend analysis complements the overall trend. From EMBSFormer-w/oRecnt, we were surprised to find that by considering only multi-periodicity, we could obtain the same superior results as in many models, whereas the near-term module could not achieve such superior results by transforming the information only on its own. In comparison, the following conclusions regarding the forecasting task can be drawn: the near-term module analyses both the trending and the cyclical. The focus can be shifted to the trending information transformation, while the multi-periodic analysis task should focus on fitting the overall trending information.

\begin{table}[h]
	\caption{Training and Inference Cost Comparison. }
	\label{efficiency study}
	\begin{center}
		\renewcommand\arraystretch{1.1}
		\begin{tabular}{c|c c}
			\toprule[1.6px]
			Dataset&\multicolumn{2}{c}{PEMS04}\\
			\hline
			Model&Training (s/epoch)&Inference (s/epoch)\\
			\hline
			ASTGCN (AAAI’19) &\underline{102.434}&\underline{26.189}\\
			GMAN (AAAI’20) &498.124&45.344\\
			ASTGNN (AAAI’19)&213.542&53.179\\
                SSTBAN (ICDE’23)&154.452&37.462\\
			EMBSFormer&\textbf{69.753}&\textbf{14.194}\\
			\toprule[1.6px]
		\end{tabular}
	\end{center}
 \vspace{-0.3cm}
\end{table}

\subsection{Efficiency Study}
\begin{figure}[h]
	\centering
	\includegraphics[width=0.8\linewidth]{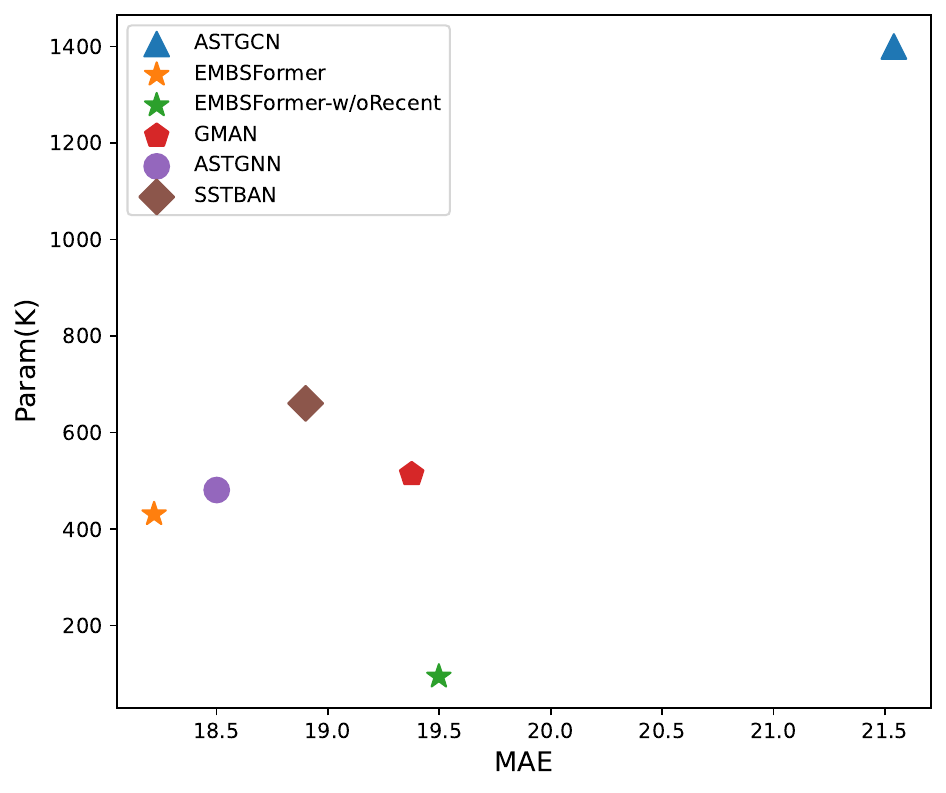}
	\caption{Parameters and Model Performance.}
	\label{params compare}
  \vspace{-0.2cm}
\end{figure}

Furthermore, we conducted an efficiency analysis comparing the EMBSFormer model with ASTGCN, ASTGNN, GMAN and SSTBAN models by training and inference time, as shown in Tab. \ref{efficiency study}. The analysis demonstrates that the EMBSFormer model exhibits excellent training efficiency in scenarios involving multiple inputs and when attention mechanisms are employed as the foundation. Compared to ASTGCN and other self-attention-based models, EMBSFormer exhibits the shortest inference time and training time. We also take into account and compare params with them, shown in Fig. \ref{params compare}. In this case, we have opted for the following parameter settings: We selected the Chebyshev polynomial order to be K=3, and the recent block num was set to 2.

In Fig. \ref{params compare}, we can observe that EMBSFormer achieves high-quality results while utilizing fewer parameters than other models. Compared to ASTGCN, which also uses multi-period(day and week), EMBSFormer utilizes fewer parameters, focuses on more information and achieves excellent accuracy. It is truly inspiring to see that the EMBSFormer-w/oRecent, which solely employs the long-term historical similarity attention mechanism module, achieves remarkable results with a parameter count of only 90K. Furthermore, it is merely 18$\%$ of the parameters used by the GMAN model, which showcases the efficiency of the long-term and short-term historical similarity attention mechanism to capture meaningful historical similarity relationships within the model.


\subsection{Case study}
In this subsection, we visually analyzed the test results on the PEMS08 dataset. As illustrated in Fig \ref{predition curve}, We visualized the training results for both 12 steps and 36 steps. The left side represents 12 steps, while the right side represents 36 steps. And the visualization time range covers from July 1, 2016, to July 3, 2016, spanning from Sunday to Monday. For the 12-step prediction results, we selected sensors 60 and 128; for the 36-step prediction, we chose sensors 35 and 76. We observe that sensors 60 and 128 exhibit similar traffic flow patterns on Sundays and weekdays, while sensors 35 and 76 demonstrate distinct traffic flow on Sundays compared to weekdays. Nevertheless, we can observe from the curves that the model performs well in predicting short-term and long-term tasks for these two scenarios. This is attributed to our consideration of the impact of weekends in the embedding layer and the effectiveness of the similar attention mechanism in analyzing similarities.

\section{Conclusions}
In this paper, we rethink the traffic flow pattern and divide it into generation and transition phases. We design a parallel multi-period flow generation module and a stacked flow transition module to model them, respectively. Specifically, the parallel multi-period generation module performs soft lookup based on historical traffic flow data from $k$ significant periods through the similarity attention mechanism. And the stacked flow transition module captures traffic transition patterns between nodes through learning information interatcion in spatial neighborhoods and implied neighborhoods. In particular, we improve learning efficiency and training/inference performance by explicit modeling of multi-period and lookup. Our model outperforms existing methods in short- and long-term traffic flow prediction on three real-world datasets. In future work, we will consider adaptive multi-period expansion and apply this method to general time series forecasting problems.

\begin{acks}
To Robert, for the bagels and explaining CMYK and color spaces.
\end{acks}

\bibliographystyle{ACM-Reference-Format}
\bibliography{ASTGFormer-cite}


\end{document}